\documentclass[10pt,twocolumn,letterpaper]{article}

\usepackage{wacv}              

\usepackage{graphicx}
\usepackage{amsmath}
\usepackage{amssymb}
\usepackage{booktabs}
\usepackage[accsupp]{axessibility}  

\newcommand{\fref}[1]{Figure~\ref{#1}}
\newcommand{\tref}[1]{Table~\ref{#1}}

\usepackage[pagebackref,breaklinks,colorlinks]{hyperref}

\usepackage[capitalize]{cleveref}
\crefname{section}{Sec.}{Secs.}
\Crefname{section}{Section}{Sections}
\Crefname{table}{Table}{Tables}
\crefname{table}{Tab.}{Tabs.}

\def\wacvPaperID{1634} 
\def\confName{WACV}
\def\confYear{2024}

\begin{document}

\title{Analyzing the Domain Shift Immunity of Deep Homography Estimation}

\author{Mingzhen Shao$^{1,4}$, Tolga Tasdizen$^{2,4}$, Sarang Joshi$^{3,4}$\\
$^1$ Kahlert School of Computing, University of Utah\\
$^2$ Department of Electrical \& Computer Engineering, University of Utah\\
$^3$ Department of Biomedical Engineering, University of Utah\\
$^4$ Scientific Computing and Imaging Institute, University of Utah\\
{\tt\small shao@cs.utah.edu, \{tolga, sjoshi\}@sci.utah.edu}
}

\maketitle

\begin{abstract}
   Homography estimation serves as a fundamental technique for image alignment in a wide array of applications. The advent of convolutional neural networks has introduced learning-based methodologies that have exhibited remarkable efficacy in this realm. Yet, the generalizability of these approaches across distinct domains remains underexplored.
   Unlike other conventional tasks, CNN-driven homography estimation models show a distinctive immunity to domain shifts, enabling seamless deployment from one dataset to another without the necessity of transfer learning.
   This study explores the resilience of a variety of deep homography estimation models to domain shifts, revealing that the network architecture itself is not a contributing factor to this remarkable adaptability.
   By closely examining the models' focal regions and subjecting input images to a variety of modifications, we confirm that the models heavily rely on local textures such as edges and corner points for homography estimation. Moreover, our analysis underscores that the domain shift immunity itself is intricately tied to the utilization of these local textures.
   \footnote{https://github.com/MingzhenShao/Homography\_estimation.git}
\end{abstract}


\section{Introduction}

 Homography is a cornerstone in computer vision, offering a geometric relationship between two images capturing the same planar surface in space. The accurate estimation of homographies between images is the first step in comprehending scene geometry. This pivotal step holds the potential to greatly enhance the performance of various vision tasks, including but not limited to multi-frame HDR imaging~\cite{Gelfand_10}, multi-frame image super resolution~\cite{Wronski_19}, burst image denoising~\cite{Liu_14}, video stabilization~\cite{Liu_13}, image/video stitching~\cite{Zaragoza_13,Guo_16}, and simultaneous localization and mapping~\cite{MurArtal_15,Zou_13}. 

 Homography estimation techniques can be categorized into two main approaches: geometric methods and deep learning-based methods. Geometric methods focus on identifying meaningful correspondences, such as points or edges, within visual data and subsequently leveraging these correspondences to compute homographies.

 The efficacy of geometric methods hinges on the precision of these identified correspondences. When correspondences are easily discernible and align well, these methods tend to yield favorable results. However, challenges arise when attempting to establish correspondences across dissimilar viewpoints, which can be particularly arduous and time-intensive under specific circumstances. In certain scenarios, the need even arises for developers to devise distinct approaches tailored to diverse scenes, factoring in viewpoint variations, lighting conditions, and image characteristics, in order to attain more accurate and robust correspondence matches.
 
 Given these challenges, there has been a growing interest in leveraging deep learning-based methods in recent years. Several models have been introduced and have demonstrated impressive levels of accuracy on their respective testing datasets ~\cite{HomographyEstimation,nguyen2017unsupervised,Nowruzi_2017_ICCV,WANG2019355,Le_2020_CVPR,zhang2020content}. However, an important aspect that has yet to be explored is the performance of these models across diverse domains. Indeed, if attaining improved accuracy with deep learning-based models necessitates tailored designs for each individual case, the broader feasibility of deploying such methods may be compromised.

 Contrary to the prevailing notion that deep learning models necessitate domain adaptation when applied across various contexts, our research reveals a remarkable observation: all deep learning-based homography estimation models display immunity to domain shifts. This insight underscores the seamless functionality of these models across diverse datasets, obviating the need for domain-specific adjustments.

 We validate this remarkable domain shift immunity by conducting a comparative assessment of one model's performance across divergent domains, achieved without necessitating any fine-tuning interventions. 
 Intriguingly, this observed domain shift immunity shares strong parallels with the attributes of geometric methods, setting it apart from conventional deep learning techniques. 
 The correlation with geometric methods raises a question: Could this immunity potentially originate from the features utilized by these models, resembling the features employed by geometric methods such as edges and corner points?

 To tackle this question, the most straightforward avenue involves identifying the precise type of feature that the model extracts from the input images. However, forging a direct and unequivocal connection between inputs and outputs in deep learning models continues to pose a significant hurdle. In our study, we adopt a widely acknowledged method to approximate feature localization, manifested through the visualization of the model's focal region.

 In contrast to many other deep learning tasks, homography can be calculated with a very limited set of correspondences. 
 However, when attempting to visually dissect the feature map alongside real-world images, the abundance of information within these images introduces a complex array of disturbances during feature analysis. In order to solve this problem, we introduce a geometrically simplified shape (GSS) dataset. This dataset serves to alleviate the density of distinct feature types—edges, corners, and colors—thus facilitating a more lucid analysis.
 Another challenge arises from the potency of deep network architectures, which can obscure distinctions of diverse features within a simple task like homography estimation. 
 To address this challenge, we introduce a specialized homography estimation network (HEN) that consists of only eight convolutional layers. This minimalist network architecture not only enables the measurement of accuracy fluctuations across various feature modifications but also averts concerns pertaining to immunity originating only from deep network structures. 

 Using the GSS dataset and HEN architecture, we successfully identify the focal region during estimation, predominantly situated within regions of local texture and encompassing edges and corners, as opposed to the more uniform regions displaying distinct colors.
 To substantiate the pivotal role of local texture in conferring the network's domain shift immunity, we undertake a comprehensive performance comparison under various texture densities and alterations. The outcomes confirm that models rely on local textures (edges and corner points) for homography estimation and demonstrate that the domain shift resilience is linked to local texture. As long as this foundational texture remains unaffected, the model consistently attains comparable prediction accuracy across diverse domains.
 
 Our contributions can be summarized as follows:\\
 \begin{enumerate}
     
     \item  We demonstrate the domain shift immunity inherent in deep homography estimation models, highlighting the autonomy of the network structure in achieving this exceptional trait.
     \item We introduce a carefully designed dataset (GSS) and architecture (HEN) tailored for dissecting the domain shift immunity exhibited by deep homography estimation models.
     \item We substantiate that the models depend on local textures for accurate homography estimation, and the observed domain shift immunity is intricately linked to the utilization of these local textures.
 \end{enumerate}

\section{Related work}

Traditional homography estimation methods typically rely on matched image feature points, such as SIFT~\cite{lowe_2004}, SURF~\cite{10.1007/11744023_32}, ORB~\cite{orb6126544}, LPM~\cite{maLPM2017}, GMS~\cite{Bian2020gms}, SOSNet~\cite{sosnet2019cvpr}, LIFT~\cite{lift2016}, and OAN~\cite{zhang2019oanet}. Once a set of corresponding features is obtained, the homography matrix is typically estimated using Direct Linear Transformation (DLT)\cite{Hartley2004} along with outlier rejection techniques such as RANSAC\cite{10.1145/358669.358692}, IRLS~\cite{03610927708827533}, and MAGSAC~\cite{8953287}.

These conventional methods heavily depend on the quality of the captured image features. When feature correspondences are accurately established, these methods tend to exhibit good performance. However, their accuracy can be compromised by an insufficient number of matched points or poor feature distribution. This limitation is often encountered in situations involving textureless regions (\eg, sky, ocean, grassland), repetitive patterns (\eg, forest, bookshelf, symmetrical buildings), or variations in illumination. 
Moreover, the presence of dynamic objects (\eg, a moving bus) further challenges the effectiveness of outlier rejection techniques. 

Another category of traditional methods for homography estimation is known as direct methods. For example, the Lucas-Kanade algorithm~\cite{1623264.1623280} is in this category, computing the sum of squared differences (SSD) between two images to guide image shifts and update the homography. An advanced technique, the enhanced correlation coefficient (ECC)~\cite{4515873}, has been proposed as a more robust replacement for SSD. In comparison to feature-point-based approaches, these direct methods are more susceptible to interference factors such as dynamic objects and variations in illumination.

 In recent years, inspired by the success of various deep learning-based methods across a range of challenging tasks, DeTone~\etal~\cite{HomographyEstimation} introduced the first deep learning-based homography estimation model. This model, comprising just eight convolutional layers, presents an end-to-end approach to homography estimation. It takes original and transformed images as inputs, using the transformation matrix as ground truth for supervised training. Subsequent studies~\cite{Nowruzi_2017_ICCV,Le_2020_CVPR} have embraced this framework, augmenting performance by substituting the backbone with more intricate network architectures.

 Supervised training, however, presents a conspicuous challenge: acquiring accurate homography matrices from real image pairs proves to be a formidable task. A popular strategy involves employing synthetic transformed images to circumvent this issue, yet this method occasionally introduces depth disparities. 
 In response, novel avenues have been explored, leading to the proposal of unsupervised training techniques. Nguyen~\etal~\cite{nguyen2017unsupervised} introduced an ingenious unsupervised approach that calculates a photometric loss between two images, coupled with the utilization of a spatial transform network (STN)\cite{NIPS2015_33ceb07b} for image warping. Building upon this approach, Zhang~\etal~\cite{zhang2020content} presented an innovative methodology for cultivating a content-aware mask, departing from the conventional approach of directly computing loss based on intensity and uniformity across the image plane. This advancement aims to bolster the prediction accuracy of unsupervised training procedures.


 Learning-based methods have demonstrated the capacity to attain pixel-level performance that greatly surpasses that of traditional approaches. Nevertheless, unlike their conventional counterparts, these methods do not explain their capability to effectively handle images from diverse domains or provide insight into the inner mechanics governing their estimation processes. 
 The concept of domain shift immunity, critical for numerous vision tasks, is closely intertwined with the deployability of a homography estimation method. Although deep learning-based approaches can offer greater accuracy, the practicality of these methods may be compromised if achieving such accuracy necessitates domain-specific fine-tuning.
 
\section{Methods}

\subsection{Reforming the homography matrix}
 The most widely used representation of a homography is a $3\times 3$ transformation matrix and a fixed scale. 
 Using $[u, v]$ for pixels in an image and $[u', v']$ for their projection onto another image in homogenous coordinates, we get the representation of a homography matrix as follows: 
    \begin{equation}
        \left( \begin{tabular}{c}
          $u'$  \\
          $v'$  \\
          $1'$  \\
        \end{tabular} \right) = 
        \left(\begin{tabular}{ccc}
             $H_{11}$ & $H_{12}$ & $H_{13}$  \\
             $H_{21}$ & $H_{22}$ & $H_{23}$  \\
             $H_{31}$ & $H_{32}$ & $H_{33}$  \\
        \end{tabular}\right)
        \left(\begin{tabular}{c}
             ${u}$ \\
             ${v}$ \\
             ${1}$ \\
        \end{tabular}\right)
    \end{equation}
 However, these nine parameters ($H_{11}, H_{12}, ..., H_{33}$) blend rotational and translational terms performed on different scales into a single vector.
 Employing these parameters directly for training deep learning models can result in imbalance issues, intensifying the training challenge.
 To address this concern, we adopt the utilization of four 2D offset vectors (comprising eight values) to represent the homography matrix.

 To derive four 2D offset vectors from a homography, we initiate the process by selecting four points denoted as $(u_i, v_i), i\in[1,4]$ to form a rectangular configuration.
 Next, we identify the corresponding four points within the homogeneous coordinate plane, represented as $(u'_i, v'_i), i\in[1,4]$.
 Subsequently, we calculate the 2D offset vectors as $\Delta u_{i} = u'_{i} - u_{i}$ and $\Delta v_{i} = v'_{i} - v_{i}$, for each $i\in[1,4]$.

 Using the four offset vectors, the process of obtaining the homography matrix $H$ with 8 degrees of freedom becomes straightforward through the solution of a linear system. The four-point parameterization manifests a homography in the following manner:

    \begin{equation}
        H_{4point} = \left( \begin{tabular}{cc}
            $\Delta u_{1}$ & $\Delta v_{1}$ \\
            $\Delta u_{2}$ & $\Delta v_{2}$ \\
            $\Delta u_{3}$ & $\Delta v_{3}$ \\
            $\Delta u_{4}$ & $\Delta v_{4}$ \\
        \end{tabular} \right)
    \end{equation}

\subsection{Homography estimation network (HEN)}
    \begin{figure}[h]
        \centering
        \includegraphics[width=0.95\linewidth]{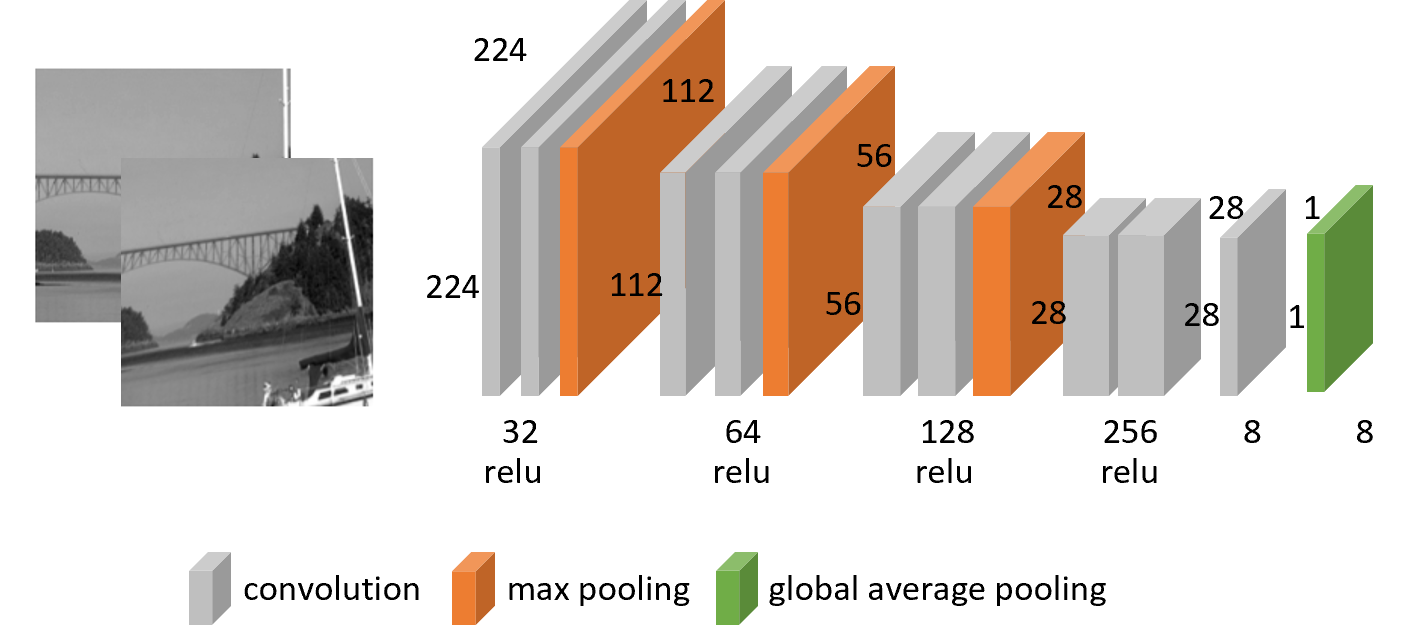}
        \caption{Homography estimation network structure.}
        \label{fig:net_structure}
    \end{figure}
    
 Several recent studies have introduced intricate models with sophisticated backbones, aiming to attain superior accuracy or robustness in comparison to the initial approach proposed by DeTone~\etal. However, the incorporation of potent backbones in these models could obscure subtle distinctions stemming from diverse input features, thereby posing a challenge when attempting to dissect the influence of different features.
 
 To effectively discern significant variations when manipulating different features, we need to employ a network that exhibits \textbf{heightened sensitivity} to feature alterations. As a result, we introduce a homography estimation network (HEN) composed of only nine convolutional layers. HEN employs a global average pooling (GAP) layer, converting the eight-channel feature maps into eight output values. Unlike conventional networks that rely on fully connected (FC) layers, HEN's utilization of the GAP layer provides better accuracy and establishes a more transparent relationship between the ultimate prediction and the feature maps.

 Our HEN operates on a grayscale original and transformed image pair as input, generating eight values ($H_{4point}$) as a representation of the homography matrix. 
 HEN may not achieve state-of-the-art accuracy due to its relatively shallow architecture, but it still delivers robust pixel-level precision, surpassing that of a classical ORB descriptor with the RANSAC method. 
 Given that classical methods are regarded as domain-unrelated, the achieved accuracy of HEN is sufficient for conducting an analysis of domain shift immunity, particularly if the observed changes in accuracy remain within a range comparable to or lesser than that of the classical methods.
 The structural layout of the proposed homography estimation network is illustrated in \fref{fig:net_structure}.

\subsection{Data generation}
    \begin{figure}[ht]
        \centering
        \begin{subfigure}[t]{.3\linewidth}
    	    \centering
    		\includegraphics[width=\linewidth]{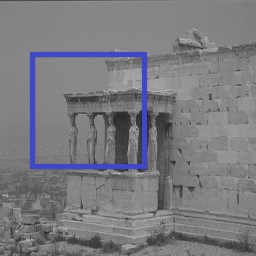}
    		\caption{Randomly crop a square at position $p$ from the original image $I$ as {\color{blue}$I_s$}.}
    		\label{fig:}
    	\end{subfigure}%
    	\hfill
    	\begin{subfigure}[t]{.3\linewidth}
    	    \centering
    		\includegraphics[width=\linewidth]{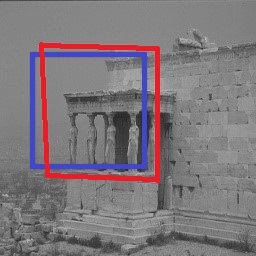}
    		\caption{Perturb four corners of the square to get a {\color{red}tetragon} and compute the homography $H$.}
    		\label{fig:}
    	\end{subfigure}%
    	\hfill
    	\begin{subfigure}[t]{.3\linewidth}
     \centering
    		\includegraphics[width=\linewidth]{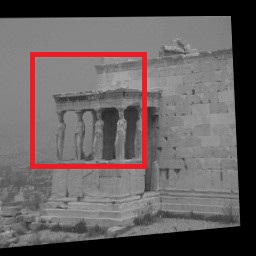}
    		\caption{Apply $H^{-1}$ to $I$ and crop a square at the same position $p$ as image {\color{red}$I_d$}.}
    		\label{fig:}
    	\end{subfigure}%

        \caption{Three steps of the data generation process.}
        \label{fig:data_generate}
    \end{figure}

 We adhere to the identical experimental setup employed in prior studies, wherein synthetic image pairs are utilized for both training and evaluating networks. This methodology affords us the capability to create image pairs spanning diverse domains while maintaining pixel-level ground truth.

 The generation process unfolds through three distinct steps, each outlined in \fref{fig:data_generate}.
 Initially, a square patch $I_s$ of dimensions $128\times 128$ is randomly cropped from the reference image $I$, positioned at $p$. To avoid subsequent bordering artifacts during the data generation pipeline, border areas are intentionally avoided.
 Subsequently, the four corners of the image patch $I_s$ undergo random perturbations by values denoted as $\delta$, within the range of $[-32, 32]$. Consequently, these four perturbed correspondences collectively define a homography $H$. Employing the inverse of this homography, $H^{-1}$, we transform the reference image $I$ into a new image denoted as $I'$.
 A second patch, $I_d$, is then cropped from $I'$ at the identical position $p$. The two patches, $I_s$ and $I_d$, are stacked channel-wise and serve as a 2-channel input for the model.

\subsection{Geometric simple shape (GSS) dataset}
 Visualizing the focus of a deep learning model is a widely employed technique for model analysis. However, homography estimation, distinct from various other deep learning tasks, has the capacity to be computed from a very limited number of correspondences. Consequently, the act of visualizing the focus of networks on commonplace datasets (\eg, BSD300, AFLW2000) could yield regions of elevated response without evident or coherent logical consistency.

 Thus, we have developed a purposeful dataset named geometric simple shape (GSS), designed to minimize potential disruptions arising from an excess of information during focus analysis. The GSS images adhere to a consistent black background and showcase only elementary geometric entities such as squares, triangles, and circles. These forms are placed randomly, each possessing varying dimensions. Moreover, the shapes can be outlined or solidly filled with distinct colors (grayscale) to assess the focus on diverse feature types, encompassing texture, color, and more.

\begin{figure}[h]
    \centering
    \includegraphics[width=0.95\linewidth]{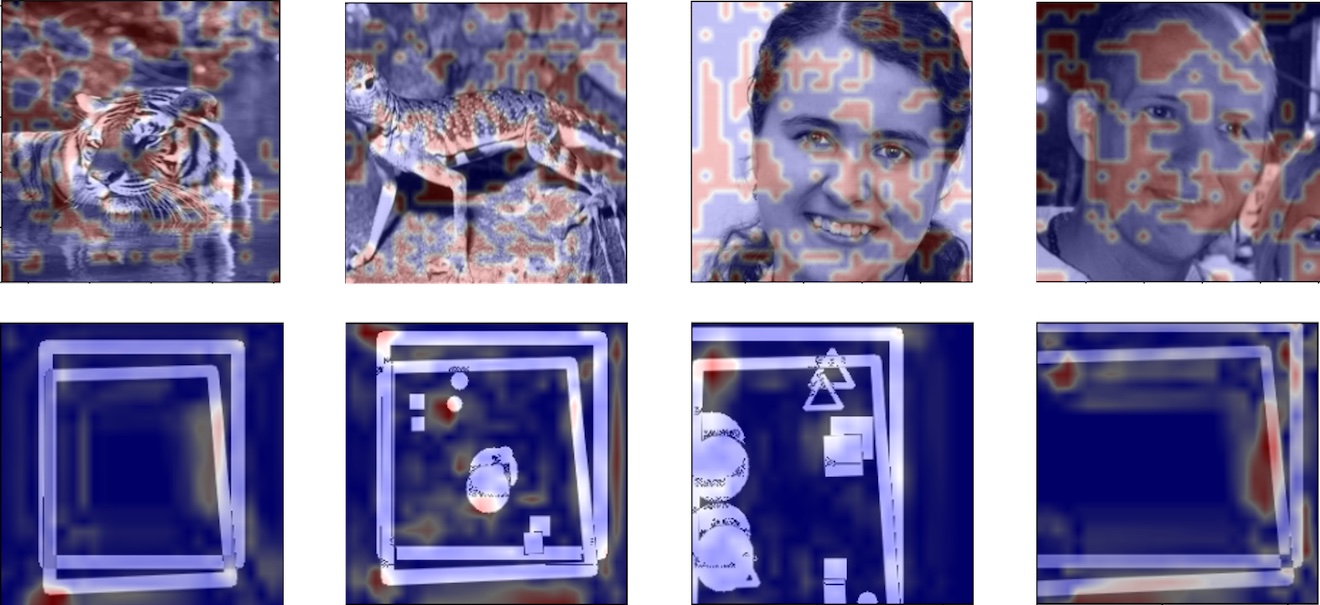}
    \caption{Visualization results of different datasets. {(top: BSD300 \& AFLW2000, bottom: GSS)}}
    \label{fig:gss}
 \end{figure}

 The focus visualizations on various datasets are illustrated in \fref{fig:gss} using the class activation mapping (CAM) technique introduced by Zhou~\etal~\cite{zhou2016learning}. By superimposing the heatmap onto the original image, we effectively highlight the areas of focus within HEN.
 This visualization approach shows that the proposed GSS dataset significantly enhances our ability to deduce the specific types of features that the network is emphasizing.

\section{Experimental results and analysis}

\begin{table*}[ht]
    \centering
    \begin{tabular}{ccccc}
        \toprule
        Model  & \multicolumn{4}{c}{Dataset}  \\ 
             & AFLW2000 & MS-COCO  & BSD300 & ISBI\\
        \midrule
        ORB+RANSAC  & 12.37  & 11.57  & 12.33  & {11.03}   \\
        ResNet50    & 1.11 & 1.15  & 1.05 & {0.77}  \\
        VGG16       & 3.42 & 3.65  & 2.60 & {2.25}  \\
        HEN         & 5.84 & 5.65  & 5.47 & {4.95} \\
        \bottomrule
    \end{tabular}
    \caption{Accuracy of predictions across diverse domains measured in MAE in pixels for various models.}
    \label{tab:compare_3}
\end{table*}

In this section, we first demonstrate the inherent domain shift immunity of deep homography estimation models. Through an evaluation across various network architectures, we establish that this immunity remains consistent regardless of the chosen network structure.

To delve into the contributing factors behind this immunity, our approach entails the initial visualization of the model's focus region during estimation using our GSS dataset. Our findings highlight a pronounced emphasis on local texture regions during the estimation process. 
Then, we conducted performance comparisons across varying local texture densities to pinpoint the pivotal features for accurate homography estimation. Our results substantiate that local texture emerges as the critical feature governing homography estimation. 

To determine the potential linkage between domain shift immunity and the utilization of local texture in homography estimation,  we test the model's performance across datasets distorted in terms of local texture.
Through these experiments, we illustrate the central role that local textures play in domain shift immunity.

In our experiments, we train all the models on the BSD300 dataset using the proposed data generation method, and subsequently test them on target datasets without any additional fine-tuning.

\subsection{Domain shift immunity}
 To demonstrate domain shift immunity, we have created a selection of multiple datasets that encompass a wide array of content domains. Our chosen datasets include BSD300, MS-COCO, AFLW2000, and ISBI, which respectively cover scenery, faces, and cells. These datasets exhibit varying levels of texture density. Specifically, the ISBI dataset displays the highest local texture density, whereas the AFLW2000 dataset includes a higher proportion of flat regions. 
 
 To establish the independence of immunity from different network structures, we proceed with a performance comparison encompassing a range of models. This comparison involves the proposed HEN network alongside more intricate and deep networks such as VGG16 and ResNet50.

 In \tref{tab:compare_3}, we first notice that even the most shallow network, HEN, effectively maintains domain shift immunity. This performance highlights that such immunity is not exclusively linked to deep architectures or confined to specific structural intricacies. Importantly, domain shift immunity doesn't mandate identical accuracy across datasets. Instead, it signifies the network's capacity to consistently deliver accurate predictions across diverse datasets. In comparison to the fluctuations observed in classical methods (ORB+RANSAC), the variability we observe in deep learning-based models does not compromise the core principle of domain shift immunity.
 
\begin{figure*}[h]
    \centering
    \includegraphics[width=0.9\linewidth]{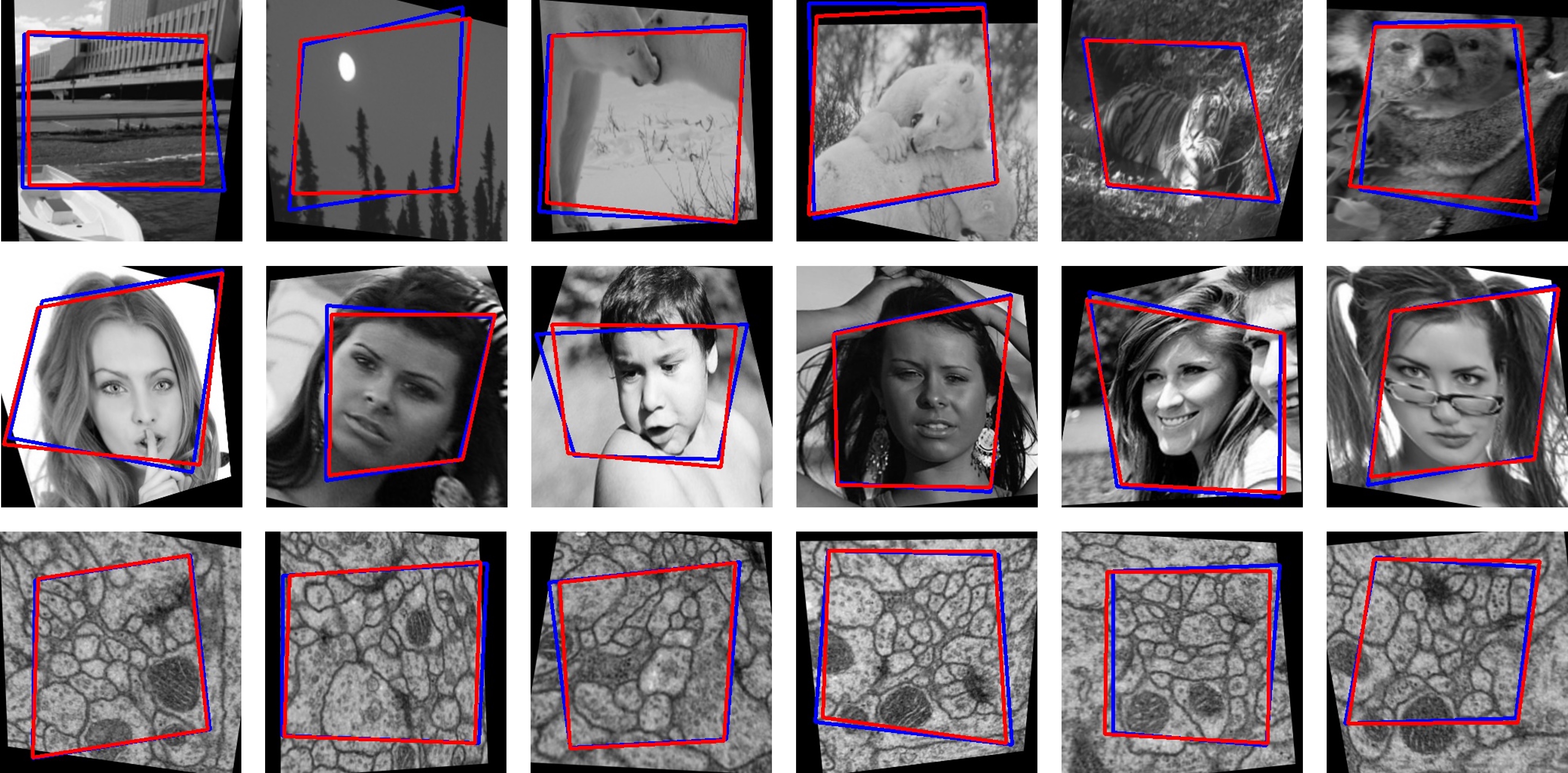}
    \caption{Results of predictions across diverse domains using HEN. The ground truth is depicted in {\color{blue}blue}, while the predictions are illustrated in {\color{red}red}. (From top to bottom: BSD300, AFLW2000, and ISBI)}
    \label{fig:data_diff_domain}
\end{figure*}

 We also observe a consistent trend in which the overall performance across all domains improves with increasing network depth. This phenomenon is easily explainable by a well-established attribute: deeper architectures have more capability to encapsulate intricate features. Notably, the sub-pixel accuracy achieved through ResNet50 reaffirms that the employment of heavier structures is superfluous for this particular task. 
 Some results of predictions using HEN are presented in \fref{fig:data_diff_domain}.


\subsection{Focus visualizations}

 The concept of domain shift immunity underscores that homography estimation models leverage specific common features present across diverse domains for accuracy.
 To identify these features, a direct approach involves visualizing the regions upon which the model focuses during estimation. However, we refrain from employing deep networks such as ResNet50 for this analysis. These powerful networks, as indicated in \tref{tab:compare_3}, are adept at extracting information from inputs, which could obscure subtle distinctions caused by alterations in specific textures. Therefore, to ensure clarity in our analysis, we choose to utilize the proposed HEN architecture in conjunction with the GSS dataset.

 Before delving into focus visualization, we conduct a straightforward experiment to demonstrate that the resultant visualization outcomes effectively pinpoint the high-contribution regions employed during estimation.
 The experiment juxtaposes the performance of two feature densities, namely \textit{normal2gap} and \textit{selected2gap}. In the \textit{normal2gap} approach, all outputs of the final convolutional layer are used as input to the GAP layer. Conversely, in the \textit{selected2gap} approach, only the top $80\%$ high-response features based on the focus map are utilized. Given that the GAP layer computes the mean of inputs, it logically follows that features containing more pertinent information for prediction would result in heightened accuracy.
 
 \begin{table}[h]
 
     \centering
    \begin{tabular}{lcc}
        \toprule
        Feature type & \textit{normal2gap} & \textit{selected2gap} \\
        \midrule
        MAE (pixel) & 12.03 & \textbf{10.37} \\
        \bottomrule
    \end{tabular}
    \caption{Prediction accuracy on GSS with varying features.}
    \label{tab:feature_type}
 \end{table}

In \tref{tab:feature_type}, we observe an approximate accuracy improvement of $1.7$ pixels when using the \textit{selected2gap} approach. This result provides evidence for the efficacy of the visualization method in identifying the regions that contribute to the estimation process.

The accuracy of predictions on the GSS dataset (\textit{normal2gap}) is slightly lower compared to other datasets such as MS-COCO and BSD300. This reduction in accuracy can be attributed to the inherent information density within the GSS dataset. The presence of extensive black (0) regions in the GSS dataset, which lack relevant information for prediction, likely contributes to this decline in accuracy. However, in the context of the performance exhibited by classic methods on the same dataset (31.43 pixels), a prediction error of around 12 pixels can still be considered reasonable.

\begin{figure*}[h]
    \centering
    \includegraphics[width=0.9\linewidth]{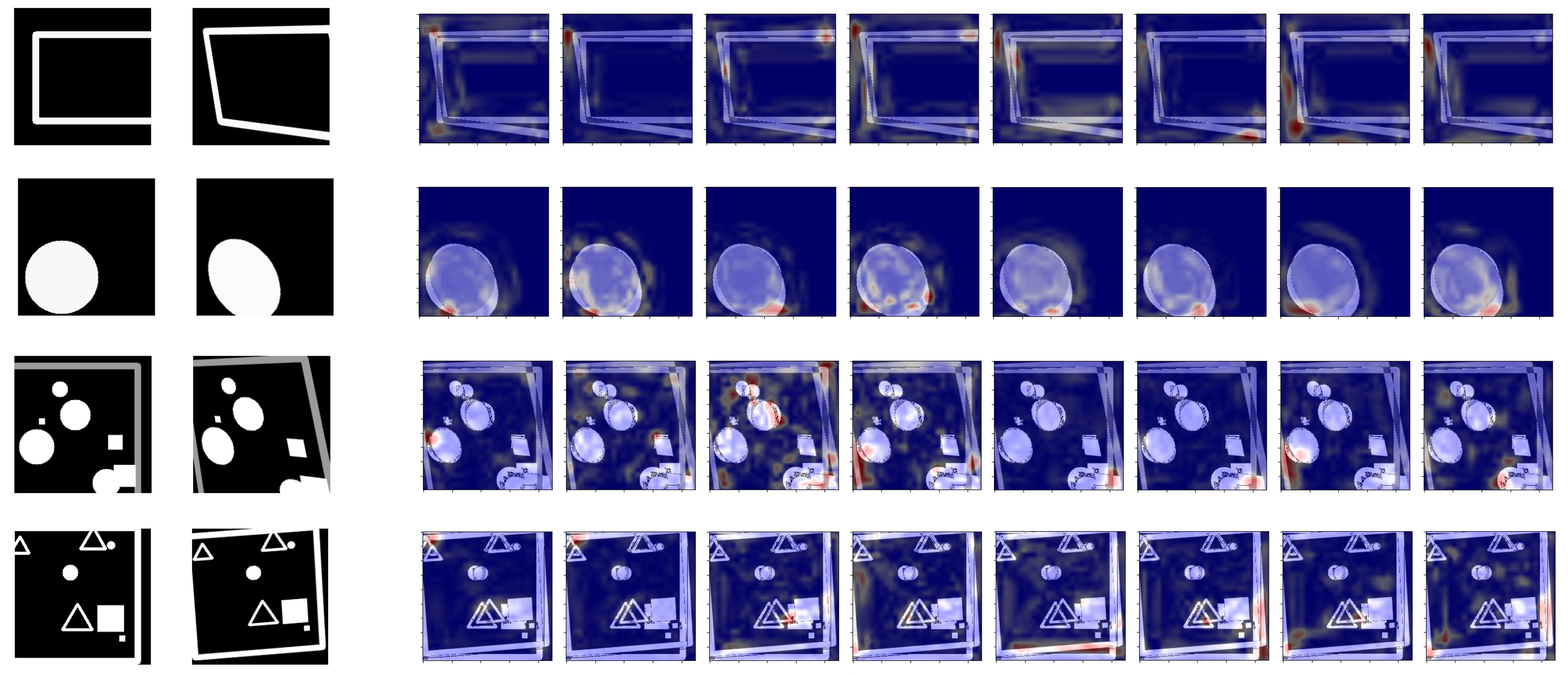}
    \caption{Visualization of focus maps on the GSS dataset. The left two columns display input image pairs, and the right eight columns depict focus visualization across output channels.}
    \label{fig:focus_GSS}
\end{figure*}

 In \fref{fig:focus_GSS}, a collection of visualization results on the GSS dataset is presented. The initial two columns show the input image pairs, whereas the subsequent eight columns exhibit the visualization of focus on the output channels. To reveal the areas of focus, the heatmap of each channel is superimposed onto the original inputs. Upon analyzing the outcomes, it becomes evident that the regions exhibiting high response are closely aligned with local texture segments, rather than regions defined by different colors (row 2 in the \fref{fig:focus_GSS}). Furthermore, given that homography estimation is achievable with minimal correspondences, not all edges evoke a significantly heightened response.

\subsection{Critical features for homography estimation}

 Expanding on the insights derived from our focus map visualizations, we found the homography estimation models' tend to prioritize regions rich in local texture during the estimation process.

 In order to furnish tangible proof of the pivotal role local textures play in homography estimation, we employ a direct approach. Our objective is to determine the significance of local textures by comparing the performance of the HEN across different texture densities.
 If deep learning models depend on local textures for accurate homography estimation, images containing a higher prevalence of such textures should yield improved performance. To validate this hypothesis, we create images with different numbers of shapes from our GSS dataset and evaluate the performance of the HEN trained on BSD300. The results of these predictions are presented in \tref{tab:acc_type}. 
 
 We observe that the input images with more shapes exhibit better performance. 
 This result confirms the critical role played by local textures in homography estimation, as the distinction between these images is based only on the density of local textures.

    \begin{table}[h]
        \centering
        \begin{tabular}{lcccccc}
            \toprule
            Number of shapes     &   1   &   5   &   9   & 15    \\
            \midrule
            MAE (pixel)      & 13.25  & 12.45  & 11.75  &  \textbf{10.23}\\
            \bottomrule
        \end{tabular}
            \caption{Prediction accuracy across various numbers of shapes.}
        \label{tab:acc_type}
    \end{table}

  \begin{figure*}[h]
     \centering
     \includegraphics[width=0.9\linewidth]{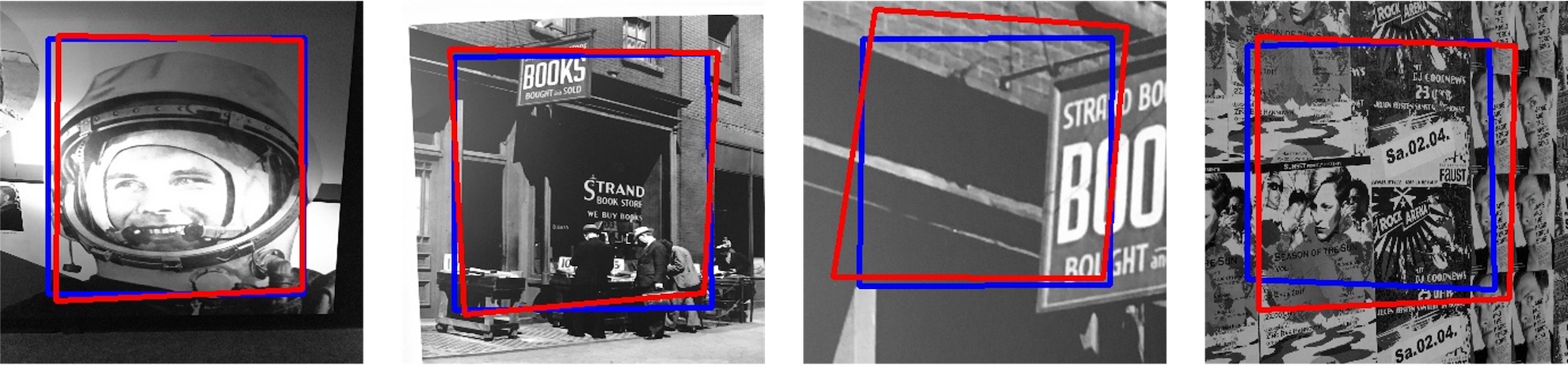}
     \caption{Validating domain shift immunity in the HPatches dataset using HEN.}
     \label{fig:phy_img}
 \end{figure*}
 
\subsection{Decoding the domain shift immunity}

 Based on our experiments, we have found that homography estimation models rely on local textures to attain accurate predictions. Remarkably, these models exhibit robust immunity to domain shift (comparable accuracy) when applied to datasets containing a diverse array of local textures. 
 However, this immunity seems to be less resilient, resulting in lower accuracy when the models process datasets characterized by lower texture density, such as the GSS dataset.

 This performance discrepancy strongly suggests that the core of domain shift immunity is linked to local textures. To substantiate this association, we undertake a series of experiments in which we apply a 3x3 Gaussian kernel to blur different datasets, thereby reducing their local textures. The ensuing performance of the identical model on these altered datasets is documented in \tref{tab:dis_acc}.

\begin{table}[h]
    \centering
    \begin{tabular}{ccccc}
        \toprule
        Model  & \multicolumn{4}{c}{Blurred Dataset}  \\ 
             & AFLW2000 & MS-COCO  & BSD300 & ISBI\\
        \midrule
        HEN         & 7.11 & 6.82  & 6.84 & {7.22} \\
        \bottomrule
    \end{tabular}
    \caption{Accuracy of HEN predictions across different blurred datasets. (measured in MAE in pixels)}
    \label{tab:dis_acc}
\end{table}

 We observe a decrease of approximately 1.5 pixels in prediction accuracy across all domains after applying the blur. This finding further validates that the domain shift immunity is primarily attributed to local textures.

 One concern regarding our explanation is that it is based on synthetic datasets. However, although physical-world image pairs may contain noise, depth information, and lighting changes, these factors do not notably affect local textures. To address this concern, we further demonstrate the persistence of domain shift immunity when applying a HEN model trained on the synthetic BSD300 dataset to real-world images from the HPatches dataset~\cite{balntas2017hpatches}, as shown in \fref{fig:phy_img}.

\section{Conclusions}
 This paper examines the domain shift immunity exhibited by deep homography estimation networks. To validate this immunity, we utilize datasets spanning diverse domains and employ various network architectures. Remarkably, our findings indicate that this immunity remains robust across various underlying network structures. 
 Through an analysis of the model's focal regions and a performance evaluation across various texture densities using the proposed dataset and architecture, we reveal the significant dependency of estimation models on local textures for precise homography estimation. Moreover, by conducting a comprehensive comparative analysis of domain shift performance under varying local texture alterations, we have established a direct correlation between the observed domain shift immunity in homography estimation networks and their fundamental reliance on the utilization of local textures.

\section{Acknowledgement}
 This work is supported by NIH Grant R01CA259686.

{\small
\bibliographystyle{ieee_fullname}
\bibliography{egbib}
}

\end{document}